%% file: main.tex
\documentclass[sigconf]{acmart}
\AtBeginDocument{%
  }

\setcopyright{acmlicensed}
\copyrightyear{2025}
\acmYear{2025}
\setcopyright{cc}
\setcctype{by}
\acmConference[MM '25]{Proceedings of the 33rd ACM International Conference
on Multimedia}{October 27--31, 2025}{Dublin, Ireland}
\acmBooktitle{Proceedings of the 33rd ACM International Conference on
Multimedia (MM '25), October 27--31, 2025, Dublin,
Ireland}\acmDOI{10.1145/3746027.3758145}
\acmISBN{979-8-4007-2035-2/2025/10}

\input{preamble}

\begin{document}

\title[Invert3D]{Align 3D Representation and Text Embedding for \\ 3D Content Personalization}


\author{Qi Song}
\orcid{0009-0006-7896-1567}

\affiliation{%
  \institution{Department of Computer Science}
  \institution{Hong Kong Baptist University}
  \state{Hong Kong SAR}
  \country{China}
}
  \authornotemark[1]
\email{qisong@life.hkbu.edu.hk}

\author{Ziyuan Luo}
\orcid{0000-0003-1580-9809}
\authornote{Equal contribution.}
\affiliation{%
  \institution{Department of Computer Science}
  \institution{Hong Kong Baptist University}
  \state{Hong Kong SAR}
  \country{China}
}
\email{ziyuanluo@life.hkbu.edu.hk}

\author{Ka Chun Cheung}
\orcid{0000-0002-2939-4686}
\affiliation{%
  \institution{NVIDIA AI Technology Center}
\institution{NVIDIA}
  \state{Hong Kong SAR}
  \country{China}
}
\email{chcheung@nvidia.com}

\author{Simon See}
\orcid{0000-0002-4958-9237}
\affiliation{%
  \institution{NVIDIA AI Technology Center}
\institution{NVIDIA}
  \country{Singapore}
}
\email{ssee@nvidia.com}

\author{Renjie Wan}
\authornote{Corresponding author. This work was carried out at the Renjie Group, Hong Kong Baptist University.}
\orcid{0000-0002-0161-0367}
\affiliation{%
  \institution{Department of Computer Science}
  \institution{Hong Kong Baptist University}
  \state{Hong Kong SAR}
  \country{China}
}
\email{renjiewan@hkbu.edu.hk}

\renewcommand{\shortauthors}{Qi et al.}

\input{sec/0_abstract}

\ccsdesc[500]{Computing methodologies~Computer vision representations}

\keywords{Embedding Alignment, 3D-to-Text, Personalized 3D Generation}



\maketitle

\input{sec/1_intro}
\input{sec/2_related_work}
\input{sec/3_method}

\input{sec/4_experiments}

\input{sec/5_conclusion}

\bibliographystyle{ACM-Reference-Format}
\bibliography{reference}


\end{document}

%% file: preamble.tex
\usepackage{amsmath,amsfonts}
\usepackage{algorithmic}
\usepackage{algorithm}
\usepackage{array}
\usepackage{textcomp}
\usepackage{stfloats}
\usepackage{url}
\usepackage{verbatim}
\usepackage{graphicx}

\usepackage{amsmath}
\usepackage{booktabs}
\usepackage{multirow}

\newcommand{\eg}{\textit{e.g.}}

\usepackage[utf8]{inputenc} 
\usepackage[T1]{fontenc}    

\usepackage{url}            
\usepackage{booktabs}       
\usepackage{amsfonts}       
\usepackage{nicefrac}       
\usepackage{microtype}      

\usepackage{epsfig}
\usepackage{graphicx}
\usepackage{amsmath}
\usepackage{multirow} 
\usepackage{caption} 

\usepackage{hyperref}       

\usepackage{cleveref}
\crefname{figure}{Fig.}{Figs.}
\crefname{table}{Table}{Tables}

\usepackage{booktabs}
\usepackage{multirow}

\usepackage{algorithm}          
\usepackage{algorithmic}

\usepackage{etoolbox}
\AtBeginEnvironment{algorithmic}{%
  \setlength{\baselineskip}{1.3\baselineskip}
  \setlength{\itemsep}{3pt}
  \setlength{\parskip}{2pt}
}

\usepackage{amsthm}        
\usepackage{amsmath}          
\usepackage{graphicx}      
\usepackage{booktabs}      
     
\usepackage{hyperref}

\usepackage{booktabs}
\usepackage{multirow}
\usepackage{makecell}
\usepackage{array}

\usepackage{balance}

%% file: sec/0_abstract.tex
\begin{abstract}
Recent advances in NeRF and 3DGS have significantly enhanced the efficiency and quality of 3D content synthesis. However, efficient personalization of generated 3D content remains a critical challenge. Current 3D personalization approaches predominantly rely on knowledge distillation-based methods, which require computationally expensive retraining procedures. To address this challenge, we propose \textbf{Invert3D}, a novel framework for convenient 3D content personalization. Nowadays, vision-language models such as CLIP enable direct image personalization through aligned vision-text embedding spaces. However, the inherent structural differences between 3D content and 2D images preclude direct application of these techniques to 3D personalization. Our approach bridges this gap by establishing alignment between 3D representations and text embedding spaces. Specifically, we develop a camera-conditioned 3D-to-text inverse mechanism that projects 3D contents into a 3D embedding aligned with text embeddings. This alignment enables efficient manipulation and personalization of 3D content through natural language prompts, eliminating the need for computationally retraining procedures. Extensive experiments demonstrate that Invert3D achieves effective personalization of 3D content. Our work is available at: \url{https://github.com/qsong2001/Invert3D}.

\end{abstract}

%% file: sec/1_intro.tex
\section{Introduction}
3D content creation has undergone a paradigm shift with the emergence of NeRF~\cite{mildenhall2020nerf} and 3DGS~\cite{kerbl2023_3dgs}, enabling high-quality scene reconstruction from multi-view images. Recent diffusion-based approaches have further simplified this process, allowing 3D generation from single images~\cite{tang2024lgm,zou2024triplane,hong2023lrm,xu2024grm,liu2023zero,liu2023one,long2024wonder3d,song2024geometry,huang2025stereo} or text prompts~\cite{shi_mvdream,lin2023magic3d,wang2023prolificdreamer,raj2023dreambooth3d}. While these advances have greatly increased the accessibility and quality of 3D content creation, efficient and flexible personalization of 3D content—especially guided by natural language—remains a largely unsolved problem.

Existing 3D personalization methods often rely on indirect operations via 2D image manipulation~\cite{wang2024gaussianeditor,chen2024gaussianeditor, yuan2022nerf_editting,haque2023instruct_nerf_editting}. Most prevalent approaches follow a multi-stage pipeline: rendering 2D images from 3D content, editing them in the 2D domain~\cite{rombach2022high,poole_dreamfusion}, and then distilling the modified information back into the 3D representation. Such knowledge distillation is computationally intensive, requiring aggregation of information across multiple viewpoints~\cite{cheng2024colorizing,song2024protecting,fan2023nerf,huang2024geometrysticker,huang2024gaussianmarker}. This process is not only time-consuming but also susceptible to information loss and geometric inconsistency, which leads to prolonged optimization cycles and suboptimal personalization quality. The limitation persists because current methods lack a direct semantic bridge between 3D representations and text embeddings~\cite{radford2021learning_clip}. As a result, they have to rely on this iterative multi-view fine-tuning procedure.

In contrast, language-guided~\cite{radford2021learning_clip,mokady2023null_text_inversion,galimage_text_inversion,parmar2023zero} personalization has achieved remarkable progress in the 2D image domain. Modern vision-language architectures—such as CLIP~\cite{radford2021learning_clip}—enable seamless image manipulation via textual prompts by establishing cross-modal alignment between visual features and textual semantics in a shared embedding space~\cite{rombach2022high, podell2023sdxl, zhang2023adding, lugmayr2022repaint,zhu2024llmbind}. This facilitates rapid and flexible content personalization through latent space operations. Establishing a similar cross-modal alignment for 3D representations would enable direct, efficient, and expressive language-guided 3D personalization. However, foundation models that can align 3D representations with textual semantics in a shared embedding space are not yet available. Unlike the 2D domain~\cite{schuhmann2022laion,liu2022taisu}, 3D-language alignment is hampered by the lack of architectural frameworks capable of native 3D processing.

\begin{figure*}[t]
    \centering
    \vspace{0.3cm}
    \includegraphics[width=1.0\linewidth]{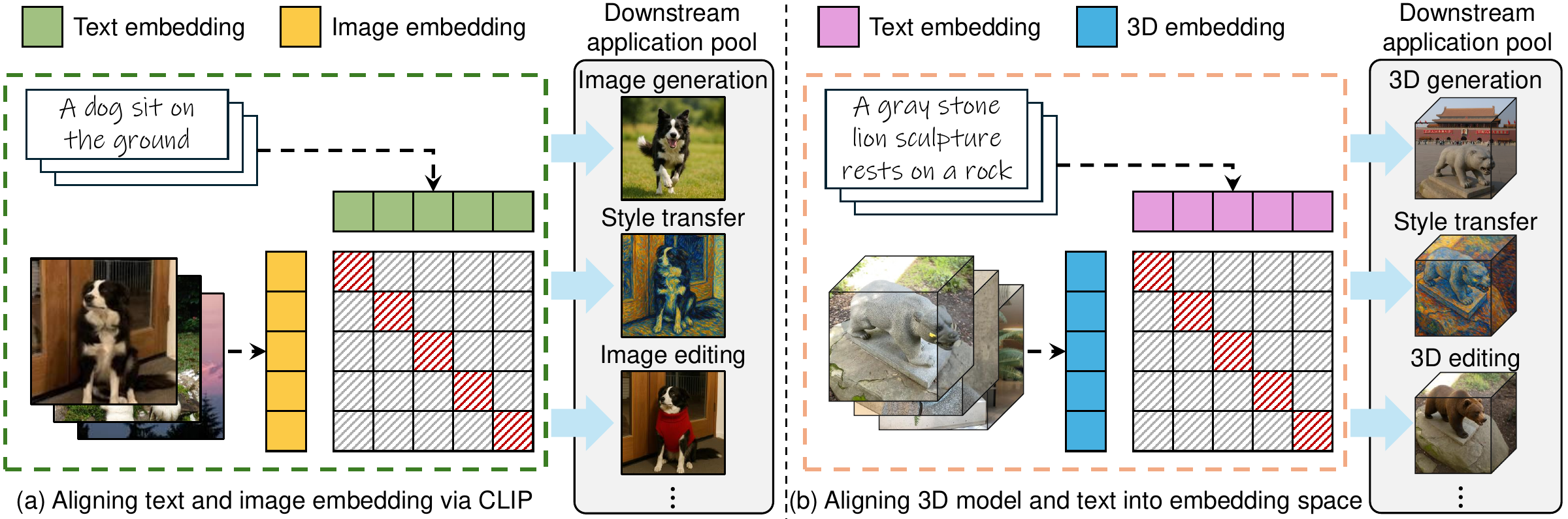}
    \caption{Overview of our envisioned scenario. (a) CLIP establishes an aligned representation of text and images within a shared embedding space, significantly advancing multimodal learning. (b) Extending cross-modal alignment to 3D and text embedding can bridge 3D representations and text in a shared semantic space. The resulting shared latent space enables intuitive language-based manipulation of 3D assets, facilitating object editing, style transfer, and other capabilities. }

    \label{fig:teaser}
\end{figure*}

We envision a novel framework for 3D content personalization without using the established multi-stage pipelines~\cite{wang2024gaussianeditor,chen2024gaussianeditor, yuan2022nerf_editting,haque2023instruct_nerf_editting,cheng2024colorizing}. As shown in Figure~\ref{fig:teaser}, we propose to invert 3D representations into a novel 3D embedding space directly aligned with text embeddings. This alignment enables direct language-driven manipulation of 3D content, allowing users to personalize 3D models by modifying input text prompts. An effective way to obtain such a embedding is to leverage inversion techniques~\cite{galimage_text_inversion,mokady2023null_text_inversion}, which have demonstrated capability in learning semantically-rich latent representations. However, existing approaches~\cite{galimage_text_inversion,mokady2023null_text_inversion} are predominantly designed for 2D images and fail to address the fundamental challenge of maintaining geometric consistency across multiple viewpoints when extended to 3D content.



To address the above difficulties, we propose \textbf{Invert3D}, a novel 3D-to-text inversion mechanism with several key innovations. \textbf{First}, we design a Camera-conditioned Text Inversion (CTI) strategy. Such a strategy uses the camera viewpoints as a part of the information during the inversion, which is able to alleviate the geometric inconsistency. Specifically, our approach continuously samples rendered 2D views from the 3D representation~\cite{mildenhall2020nerf,kerbl2023_3dgs} across diverse camera perspectives. For each sampled view, we extract its corresponding camera pose parameters and feed both the rendered image and its pose information into the inversion process of a pre-trained 2D text-to-image diffusion model~\cite{galimage_text_inversion,mokady2023null_text_inversion}. This process inverts the joint input (image + pose) into a 3D embedding aligned with the text embedding.
\textbf{Second}, to enable precise and potent control over the personalized 3D content generated using our derived 3D embedding, we introduce an attention re-weighting module. While the aligned embedding space allows initial manipulation via text prompts, we observe that simply replacing or adding new text instructions often results in subtle or incomplete personalization. Our module modulates the attention scores associated with specific semantic concepts within the text prompt during the text-to-3D generation process.

As shown in \cref{fig:teaser}, we establish a direct semantic bridge between 3D representations and text embedding. The inversion mechanism effectively encodes the 3D scene into a versatile, text-aligned embedding. This embedding captures the scene's essential structure and appearance. Subsequently, the attention re-weighting module dramatically enhance control over personalization by amplifying the impact of text instructions during generation. We demonstrate the application of aligning text-3D pairs for guided personalized 3D generations with text prompts~\cite{galimage_text_inversion,hertzprompt-to-prompt}. The contributions of this paper can be summarized as follows:

\begin{itemize}
\item We propose a novel framework for aligning 3D representations (e.g., NeRF, 3DGS) with semantic text embeddings in vision-language latent spaces. This bridges the gap between 3D representation and human-interpretable tokens, enabling direct semantic manipulation of 3D content.
\item  We demonstrate the potential of this alignment for 3D reconstruction and personalization, showing how inverse embeddings enable intuitive text-guided personalization without regenerating or tuning 3D content.
\item  We design a 3D-to-text inversion optimization that allows embeddings to effectively represent 3D contents. This direct mapping eliminates the need for iterative cross-modal distillation~\cite{poole_dreamfusion}, thereby reducing optimization costs and ensuring semantic consistency across multiple views.
\end{itemize}
We demonstrate the effectiveness of our method in both 3D reconstruction and personalized 3D generation. Extensive experiments are conducted to validate the application of our approach.

%% file: sec/2_related_work.tex
\section{Related work}

\subsection{Text-guided 3D generation and editing}
The pursuit of high-fidelity and efficient 3D scene representation drives remarkable advancements. Neural Radiance Fields (NeRF)~\cite{mildenhall2020nerf} pioneers the implicit approach by using MLPs to represent scenes as continuous volumetric functions. Subsequent studies~\cite{muller2022instant, barron2021mip, barron2022mip, luo2023copyrnerf, luo2025nerf, luo2024imaging} significantly accelerate training and improve both reconstruction quality and model security, thereby broadening NeRF's applicability. Simultaneously, explicit methods like 3D Gaussian Splatting (3DGS)~\cite{kerbl20233d} emerge as a powerful alternative. 3DGS utilizes optimized anisotropic Gaussians to represent scenes, which enables real-time rendering with exceptional visual fidelity. This marks a substantial leap forward in both efficiency and visual quality compared to previous approaches. Building upon the success of 3DGS, several subsequent works have focused on further enhancing its performance. For instance, Mip-Splatting~\cite{yu2024mip} introduces mip-mapping techniques to improve level-of-detail rendering and memory efficiency, while maintaining high visual fidelity. Other research efforts have targeted faster optimization~\cite{liu2024citygaussian,liu2025citygaussianv2}, better scalability for large-scale scenes, and improved handling of dynamic content~\cite{wu20244d,ren2023dreamgaussian4d}. 

To enable creative control, recent efforts turn to \emph{text-guided 3D generation and editing}. Techniques like DreamFusion~\cite{poole_dreamfusion} and Prolific Dreamer~\cite{wang2023prolificdreamer} leverage pretrained 2D diffusion models (e.g., Stable Diffusion) via Score Distillation Sampling (SDS), distilling 2D priors into 3D representations through iterative optimization across multiple views. Similarly, editing methods such as Instruct-NeRF2NeRF~\cite{haque2023instruct_nerf_editting} use textual instructions to iteratively update a NeRF by distilling edits from an image diffusion model applied to rendered views. ColorNeRF~\cite{cheng2024colorizing} tackles colorizing NeRFs reconstructed from multi-view monochromatic images by predicting missing color channels in Lab space with a query-based colorization strategy. For 3D Gaussian Splatting (3DGS), recent advances enable efficient text-guided editing. GaussianEditor~\cite{chen2024gaussianeditor,wang2024gaussianeditor} decomposes editing into semantic region localization and diffusion-guided attribute optimization, completing scene modifications within 20 minutes. Point’n Move~\cite{huang2024point} supports interactive rigid transformations via self-prompted 3D mask propagation, while Gaussian Grouping~\cite{ye2024gaussian} combines 2D segmentation from SAM with 3D spatial constraints for object-level manipulations. Prometheus~\cite{yang2025prometheus} further introduces a feed-forward framework with RGB-D latent space, generating view-consistent 3D Gaussians without iterative optimization.

\subsection{Vision-language embeddings alignment}

The advent of vision-language models\cite{radford2021learning_clip} has revolutionized the semantic alignment between images and text. Notably, CLIP~\cite{radford2021learning_clip} introduce large-scale contrastive pre-training on vast image-text datasets. This approach learns a shared multimodal embedding space, where semantically related images and texts are closely aligned. Such vision-language embedding alignment enables diverse and impactful applications in multi-modal learning and understanding. Key capabilities include cross-modal retrieval (finding images from text queries and vice versa)~\cite{li2022blip, li2021align}, zero-shot image classification across extensive datasets without task-specific training~\cite{sammaniinterpreting}, and vision-language understanding such as visual question answering (VQA)~\cite{schwenk2022okvqa, song2022clip} and image captioning~\cite{li2023blip, yu2023cgt}.

Building upon this foundation, diffusion models like Stable Diffusion~\cite{rombach2022high} leverage the semantic expressiveness of such aligned embeddings to generate high-fidelity images directly from text prompts. Crucially, these models operate within a shared latent space bridging vision and language, enabling intuitive image generation and manipulation via textual instructions. Techniques have emerged to leverage this aligned space for controllable editing. Textual Inversion~\cite{galimage_text_inversion, mokady2023null_text_inversion} embeds novel concepts (e.g., objects, styles) into the model's latent space from a few examples, enabling personalized generation with custom concepts in prompts. Complementarily, Prompt-to-Prompt~\cite{hertzprompt-to-prompt} facilitates attribute editing (e.g., changing ``dog'' to ``cat'') by manipulating the connection between text tokens and image regions derived from the shared embedding space. These approaches demonstrate the power of vision-language aligned embeddings for intuitive and flexible image synthesis and control. Inspired by these 2D techniques, our work proposes their adaptation to guide text-to-3D generation.

\begin{figure*}[t]
    \centering
    \includegraphics[width=1.0\linewidth]{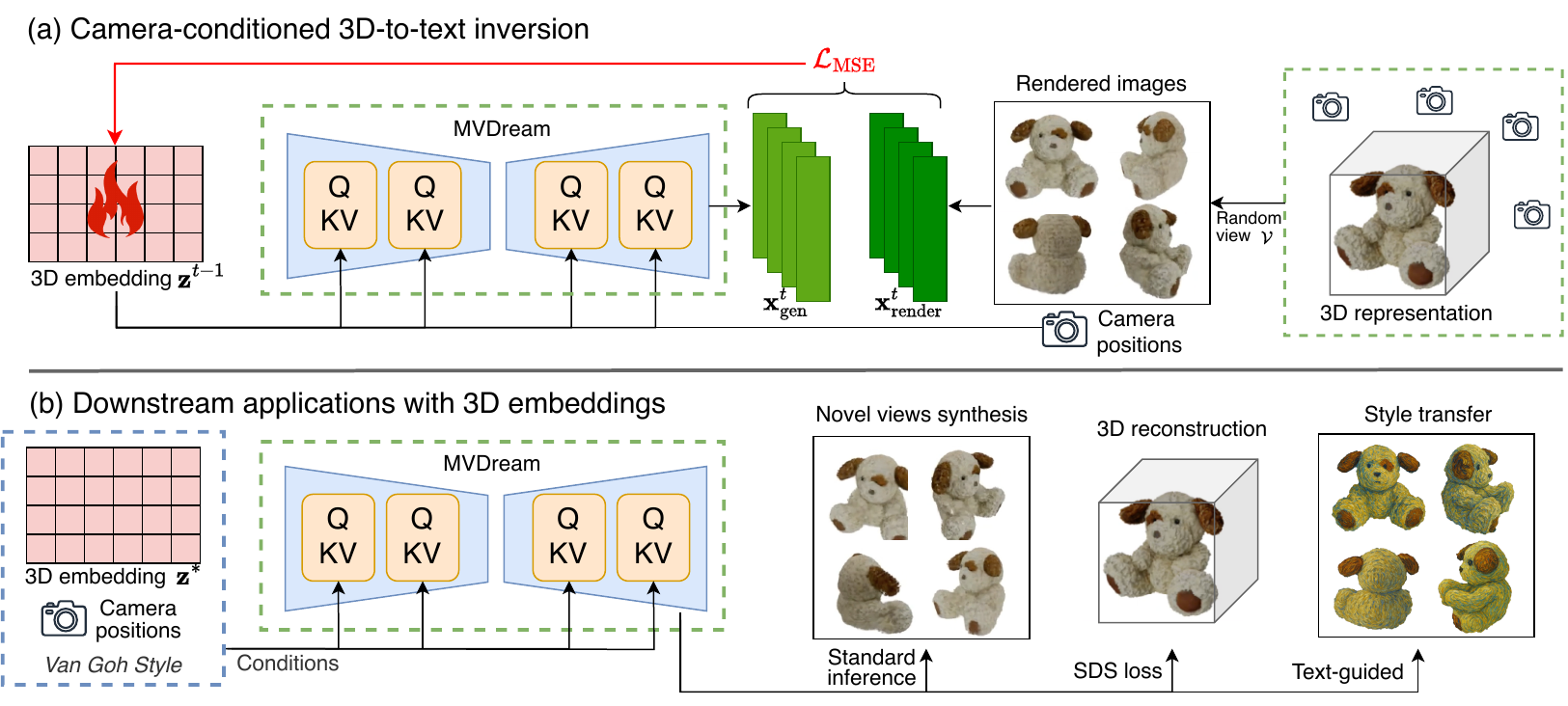}
    \caption{Our Invert3D framework. (a) We introduce a camera-conditioned 3D-to-text to obtain the 3D embedding $\mathbf{z}^*$ via camera-conditioned text inversion. (b) The obtained 3D embedding $\mathbf{z}^*$ enables reconstruction and text-guided personalization with the text-to-3D process~\cite{shi_mvdream}, including novel views synthesis, 3D reconstruction, or style transfer.}
    \label{fig:framework}
\end{figure*}

\section{Preliminary}

\subsection{Text inversion: bridging images and text}
Text inversion~\cite{galimage_text_inversion,mokady2023null_text_inversion} refers to the process of mapping a given image (or a set of images) into a point in the vision-language embedding space—typically the same space used by large vision-language models such as CLIP~\cite{radford2021learning_clip} or Stable Diffusion~\cite{rombach2022high}. This embedding captures the semantic content of the image and then provides an effective way to manipulate the image within the embedding space using a text prompt.

Formally, given an image $\mathbf{x}$, text inversion seeks to find a pseudo-word embedding $\mathbf{z}^*$ such that $\mathbf{z}^*$ can be incorporated into textual prompts. This allows the generative model (e.g., Stable Diffusion) to faithfully reconstruct $\mathbf{x}$ or produce semantically consistent variations. This process is typically achieved by optimizing $\mathbf{z}^*$ to minimize the difference between the generated image and the original image, as measured by a perceptual loss. The optimization can be expressed as:
\begin{equation}
\mathbf{z}^* = \arg\min_{\mathbf{z}} \; \mathcal{L}(G(\mathbf{z}), \mathbf{x}),
\end{equation}
where $G(\cdot)$ denotes the generative model~\cite{rombach2022high} conditioned on the embedding $\mathbf{z}$, and $\mathcal{L}$ is MSE loss function.

Recent works such as Textual Inversion~\cite{galimage_text_inversion} and Null-Text Inversion~\cite{mokady2023null_text_inversion} have demonstrated that this approach enables the embedding of novel concepts—such as specific objects, styles, or even individual identities—directly into the text embedding space. Once an image is mapped to a text embedding, it can be easily manipulated by editing the text prompt, enabling powerful and flexible downstream applications.

\subsection{Towards 3D-to-text inversion}

Inspired by the success of text inversion in the 2D domain, our goal is to extend this paradigm to the personalization of 3D content. Specifically, we aim to develop a 3D-to-text inversion mechanism that maps a 3D representation (e.g., NeRF or 3DGS) into a text-aligned embedding space. This embedding should enable direct manipulation and personalization through text prompts—without the need for iterative, view-by-view optimization.

Achieving this goal has several difficulties. 3D representations usually contain information from multiple viewpoints. Simply using the off-the-shelf text inversion method cannot effectively capture the multiview information, resulting in inconsistent results. To overcome these challenges, we propose \textbf{Camera-conditioned 3D-to-text inversion}. Our approach samples a diverse set of 2D renderings from the 3D scene under various camera poses. These renderings are then individually inverted into the text embedding space using established 2D text inversion techniques. By aggregating these per-view embeddings with camera pose conditioning, our approach constructs a unified 3D embedding that faithfully represents the entire scene in the vision-language space. Thereby, we establish the foundation for direct language-driven 3D content creation and personalization.

%% file: sec/3_method.tex
\section{Proposed method}
\label{sec:method}
Existing approaches~\cite{galimage_text_inversion,mokady2023null_text_inversion} are predominantly designed for 2D images and fail to address the fundamental challenge of maintaining geometric consistency across multiple viewpoints when extended to 3D content. Moreover, current 3D representations (\eg, NeRF~\cite{mildenhall2020nerf} and 3DGS~\cite{kerbl2023_3dgs}) are inherently complex and decoupled from human-interpretable concepts, which restricts intuitive interaction and editing.

We bridge this gap by developing the first method to embed 3D scenes directly into the vision-language semantic space~\cite{radford2021learning_clip}. To overcome the semantic-geometric mismatch in current 3D systems, we introduce \textbf{Invert3D}, a pipeline that aligns neural representations with semantic embeddings. As shown in \cref{fig:framework}, our framework transforms 3D content into editable tokens that respond to natural language. We then present how to leverage these 3D embeddings for downstream tasks, including 3D reconstruction and personalized text-guided generation. This alignment enables direct personalization of 3D scenes through a natural language prompt.

\subsection{Camera-conditioned 3D-to-text inversion}
\label{subsec:inversion}
Text inversion~\cite{galimage_text_inversion,mokady2023null_text_inversion} aims to learn a text embedding that captures the unique characteristics of a given subject. By inverting the images into the latent space, such a strategy is able to personalize the given subjects by following the information within the text prompts. These approaches typically operate on 2D images by using vision-language models~\cite{radford2021learning_clip,li2022blip}. However, standard text inversion methods fail on 3D content due to multi‐view inconsistency.  We solve this by introducing camera conditioning during inversion, ensuring the embedded representation preserves 3D structure across viewpoints. Crucially, we optimize distances in the latent space, where images are encoded as $\mathbf{x} = E(\mathbf{I})$ using Stable Diffusion's encoder $E$~\cite{kingma2013auto}.

Our Camera-conditioned Text Inversion (CTI) incorporates the camera viewpoint as an explicit input during embedding, allowing the model to account for the position and orientation from which each image is rendered. By conditioning on camera parameters, the resulting embeddings are encouraged to capture the underlying 3D structure and remain consistent across diﬀerent viewpoints. Given a 3D scene $\mathcal{S}$ represented as Gaussian splats~\cite{kerbl2023_3dgs} or neural radiance fields~\cite{mildenhall2020nerf}, we aim to obtain a 3D embedding $\mathbf{z}^*$ which can represent the 3D scene. The optimization objective is reformulated to operate in the semantics latent space:
\begin{equation}
\label{eq:inversion}
\mathbf{z}^* = \mathop{\arg\min}\limits_{\mathbf{z}} \sum\limits_{\mathbf{v} \in \mathcal{V}} \mathcal{L}_{\text{MSE}} \Big( E(\pi(\mathcal{S}, \mathbf{v})),  \mathcal{D}_{\text{latent}}( \mathbf{z}, \mathbf{c}_{\mathbf{v}} ) \Big),
\end{equation}
where $\mathcal{V}$ denotes camera poses spanning azimuth and elevation angles, and $\pi(\cdot)$ represents the differentiable rasterizer used in 3DGS~\cite{kerbl2023_3dgs} or the volumetric renderer in NeRF~\cite{mildenhall2020nerf}.
$E(\cdot)$ is the VAE encoder~\cite{kingma2013auto}, $\mathcal{D}_{\text{latent}}(\cdot)$ denotes the latent representation. In our setting, we use MVDream to produce such latent representations, and $\mathcal{L}_{\text{MSE}}$ computes the latent-space reconstruction loss.

The optimization process iteratively updates the 3D embedding. At iteration $t$, we first sample and render the views from the neural representation randomly. A camera pose $\mathbf{v}_t$ is sampled and the corresponding view is rendered via
\begin{equation}
\mathbf{I}_{\text{render}}^t = \pi(\mathcal{S}, \mathbf{v}_t).
\end{equation}
Then, the rendered view is encoded into the latent space using Stable Diffusion's encoder:
\begin{equation}
\mathbf{x}_{\text{render}}^t = E(\mathbf{I}_{\text{render}}^t).
\end{equation}

MVDream denoise the noisy latent representation conditioned on the current 3D embedding and camera parameters:
\begin{equation}
\mathbf{x}_{\text{gen}}^t = \mathcal{D}_{\text{latent}}(\mathbf{z}_t, \mathbf{c}_{\mathbf{v}_t}).
\end{equation}

The 3D embedding is refined via gradient descent to minimize reconstruction error between generated view latent $\mathbf{x}_{\text{gen}}^t$ and rendered $\mathbf{x}_{\text{render}}^t$ via:
\begin{equation}
\mathbf{z}_{t+1} = \mathbf{z}_t - \eta \nabla_{\mathbf{z}} \mathcal{L}_{\text{MSE}}(\mathbf{x}_{\text{render}}^t, \mathbf{x}_{\text{gen}}^t).
\end{equation}
This camera-conditioned formulation ensures the inverted embedding $\mathbf{z}^*$ encapsulates multi-view consistent semantics while residing natively in the latent space.

\subsection{Downstream applications}
\label{subsec:downstream}
After the camera-conditioned inversion, we obtain the 3D embeddings that represent the 3D contents. The aligned 3D embeddings can be used for 3D reconstruction and intuitive personalization capabilities that were previously unattainable.

\subsubsection{3D Reconstruction from inverted embeddings}
Our framework achieves direct 3D synthesis from vision-language embeddings by leveraging the semantic alignment established through inversion. Unlike traditional reconstruction that relies on geometric optimization, we bypass this process entirely by utilizing MVDream's~\cite{shi_mvdream} multi-view generation capability driven by the inverted embedding $\mathbf{z}^*$.

Our embedding-driven approach bypasses this by treating 3D embeddings as semantic blueprints for direct 3D synthesis. The inverted embedding $\mathbf{z}^*$ - which semantically encodes the 3D scene in vision-language space - is fed directly into MVDream:
\begin{equation}
\{\mathbf{I}_{\mathbf{v}}\}_{\mathbf{v}\in\mathcal{V}} = \mathcal{D}(\mathbf{z}^*, \{\mathbf{c}_{\mathbf{v}}\}),
\end{equation}
where MVDream~\cite{shi_mvdream} generates consistent multi-view images $\{\mathbf{I}_{\mathbf{v}}\}$ for camera poses $\mathcal{V}$.
This embedding-driven reconstruction bypasses computationally expensive SDS distillation by directly leveraging semantic priors encoded in $\mathbf{z}^*$.

\begin{figure*}[!t]
  \includegraphics[width=0.96\textwidth]{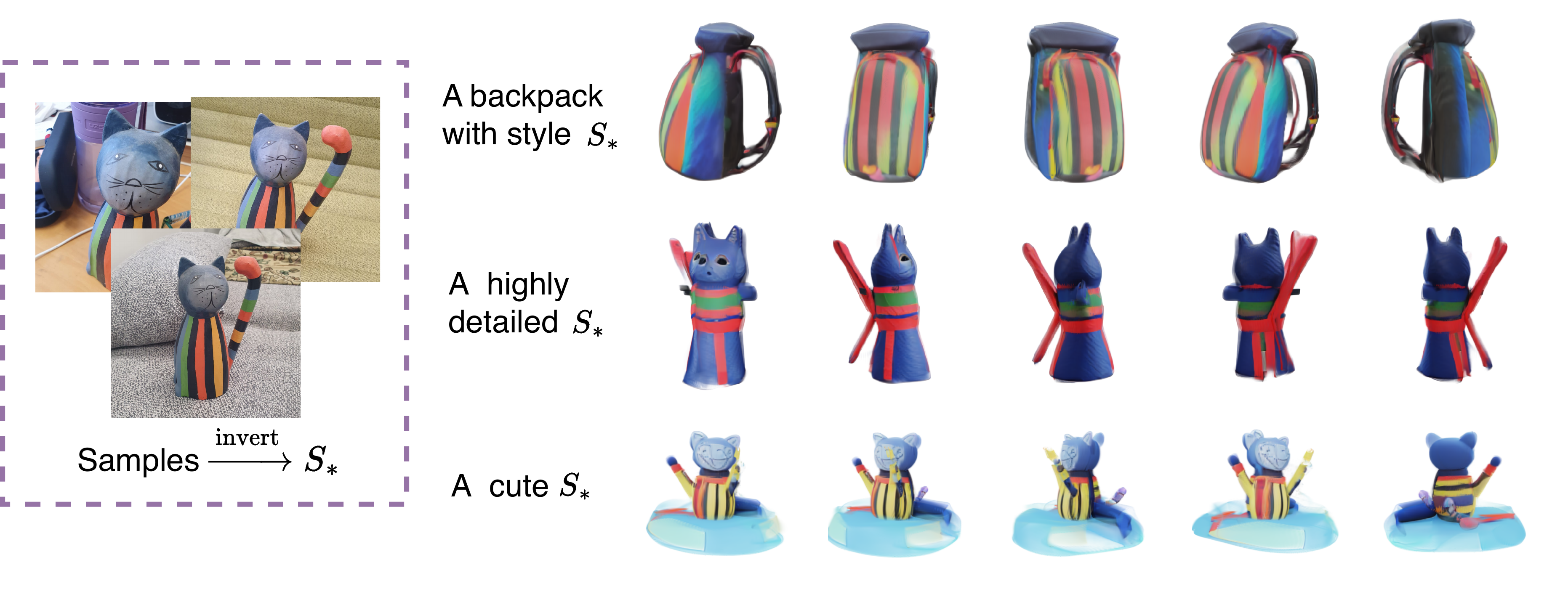}
    \caption{Personalized 3D reconstruction results obtained by aligning single‐view images with text embeddings. The inverted $S_{*}$ can be used for text-guided 3D reconstruction as it can represent the original samples~\cite{galimage_text_inversion}.}
    \label{fig:single_text}
\end{figure*}

Given an optimized 3D embedding $\mathbf{z}^*$ and the corresponding camera position, we can reconstruct 3D scenes through a standard SDS optimization for the 3D model. The reconstruction objective minimizes multi-view discrepancy between renders and MVDream outputs:
\begin{equation}
\nabla_{\mathcal{S}}\mathcal{L}_{\text{SDS}} = \mathbb{E}_{t,\epsilon} [ w(t) ( \underbrace{\hat{\epsilon}_{\phi}(\mathbf{I}_{\text{render}}|\mathbf{z}^*,t)}_{\text{guided denoising}} - \epsilon ) \frac{\partial \mathbf{I}_{\text{render}}}{\partial \mathcal{S}} ],
\end{equation}
where $\mathbf{I}_{\text{render}} = \pi(\mathcal{S}, \mathbf{v})$. By combining SDS-based methods~\cite{poole_dreamfusion}, our 3D embedding could provide a precise gradient, thereby enabling a precise 3D reconstruction.

\subsubsection{Personalized text-guided generation}

By using our 3D embedding, the user can personalize the 3D contents as easily as editing text. Our semantic alignment enables natural language operations through simple vector arithmetic in embedding space.

We enable intuitive 3D editing through vector arithmetic in the CLIP embedding space. Given source text describing current attributes and target text specifying desired modifications, we compute the semantic delta:
\begin{equation}
\begin{aligned}
\Delta \mathbf{z}_{\text{text}} &= E(\texttt{"target attribute"}) \\
&\quad - E(\texttt{"source attribute"}),
\end{aligned}
\end{equation}
where $E(\cdot)$ denotes CLIP's text encoder~\cite{radford2021learning_clip}. The edited embedding is obtained through a linear combination:
\begin{equation}
\label{eq:editing}
\mathbf{z}_{\text{edit}} = \mathbf{z}^* + \lambda \Delta \mathbf{z}_{\text{text}},
\end{equation}
where $\lambda$ controls editing intensity.  However, when editing styles like ``Van Gogh style'', we observed the inverted 3D embedding $\mathbf{z}^*$ occupies substantial capacity in the vision-language latent space, limiting the expression of new attributes. To resolve this, we adopt a re-weighting attention manipulation technique introduced from Prompt-to-Prompt~\cite{hertzprompt-to-prompt} during denoising latent in MVDream~\cite{shi_mvdream}. Let $\mathcal{P} = \{w_1, \dots, w_k\}$ be the text tokens corresponding to the added style (e.g., $w_{\text{style}} = \text{"Van Gogh"}$), and let $A^{(\ell,t)}$ be the cross-attention map at layer $\ell$ and timestep $t$. We reinitialize and amplify the attention weights for the style token:

\begin{equation}
\widetilde{A}^{(\ell,t)}_{i,j} = 
\begin{cases} 
c \cdot A^{(\ell,t)}_{i,j} & \text{if } w_j = w_{\text{style}} \\
A^{(\ell,t)}_{i,j} & \text{otherwise}
\end{cases},
\label{eq:prompt}
\end{equation}
where amplification factor $c > 1$ enhances stylistic influence. This modified attention map $\widetilde{A}^{(\ell,t)}$ replaces the original in subsequent diffusion steps. This adjustment ensures enhanced stylistic transfer while preserving geometric integrity. Feeding $\mathbf{z}_{\text{edit}}$ with modified attention into MVDream generates 3D scenes with attribute modifications consistently applied across all views.

\subsection{Implementation details}

Our framework is built upon the MVDream backbone, adopting a multi-view latent diffusion architecture for 3D-to-text inversion and generation. The model leverages a UNet-based multi-view diffusion network with 320 base channels and 8 attention heads, integrating camera pose conditioning (16 dimensions) directly into the generative process. We use a pre-trained VAE encoder with a 4-dimensional latent embedding and a resolution of $256 \times 256$ to encode rendered views. Semantic conditioning is realized via a Frozen CLIP text encoder~\cite{radford2021learning_clip}, which provides robust vision-language alignment. For personalization, we employ a learnable embedding manager that initializes tokens with the word ``object'' and allocates 32 vectors per token for increased semantic capacity.

During training, we use a batch size of 1 and sample rendered views at $256 \times 256$ resolution from 3D Gaussian Splatting or NeRF representations. 
For each 3D scene, either represented as NeRF~\cite{mildenhall2020nerf} or 3D Gaussian Splatting~\cite{kerbl2023_3dgs}, we randomly sample 4 camera poses covering diverse azimuth and elevation angles to ensure multi-view semantic consistency. The model is optimized for 6100 steps using Adam~\cite{kingma2014adam} with a base learning rate of $5\times10^{-3}$, linearly scheduled between $0.00085$ and $0.012$ over 1000 diffusion timesteps. To encourage multi-view consistency and semantic fidelity, camera parameters are concatenated with the latent features during both inversion and synthesis. For text-guided personalization, cross-attention re-weighting~\cite{hertzprompt-to-prompt} is activated within the UNet's transformer blocks, amplifying the influence of style or attribute tokens during generation.

%% file: sec/4_experiments.tex
\begin{figure*}[t]
    \centering
    \includegraphics[width=0.90\textwidth]{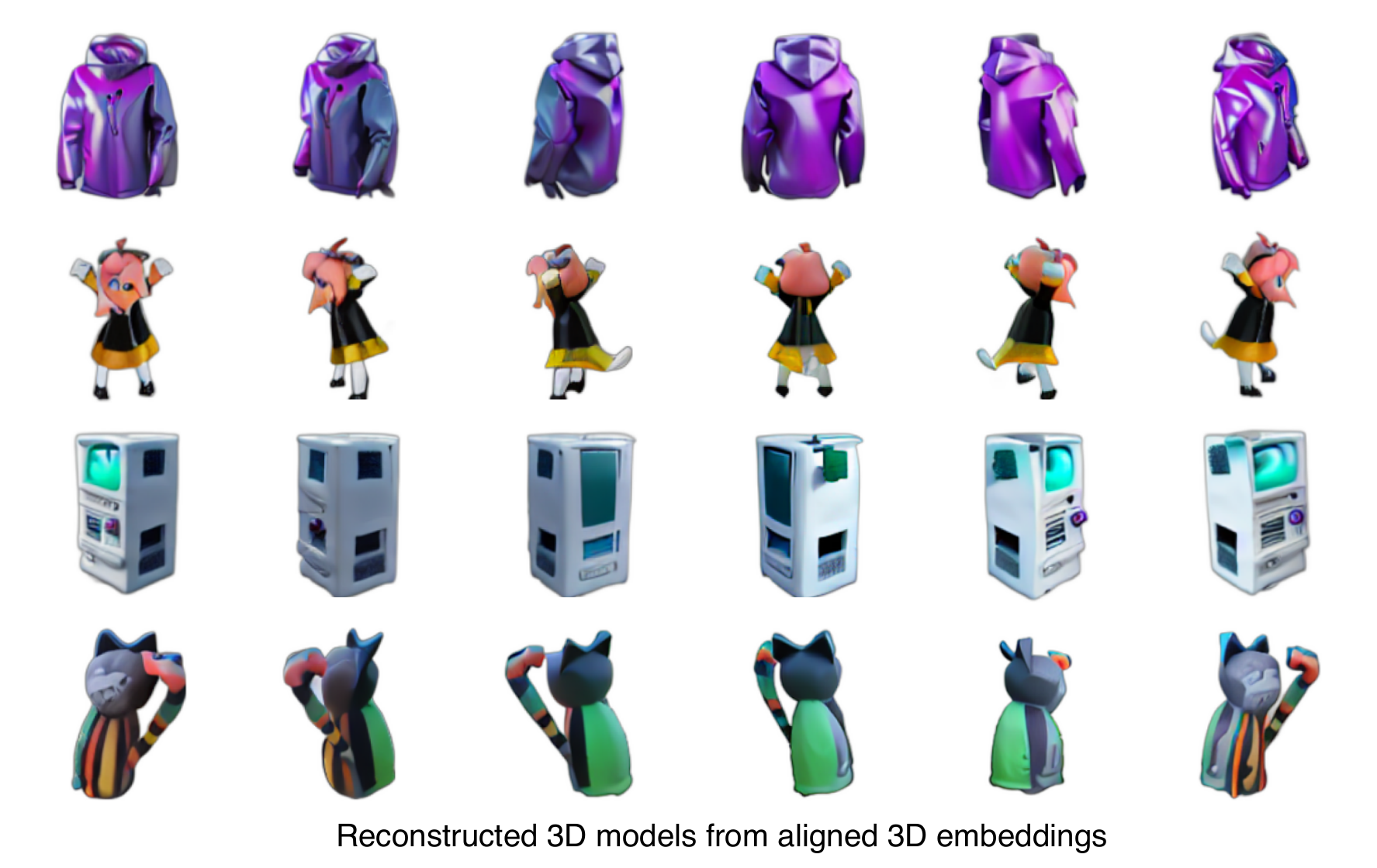}
    \caption{Visualized results of 3D reconstruction from aligned 3D embeddings. These novel views are directly generated with 3D embeddings, demonstrating their effectiveness in representing the original 3D contents.}
    \label{fig:3d_rec}
\end{figure*}

\begin{figure*}[!t]
  \centering
  \includegraphics[width=1.00\textwidth]{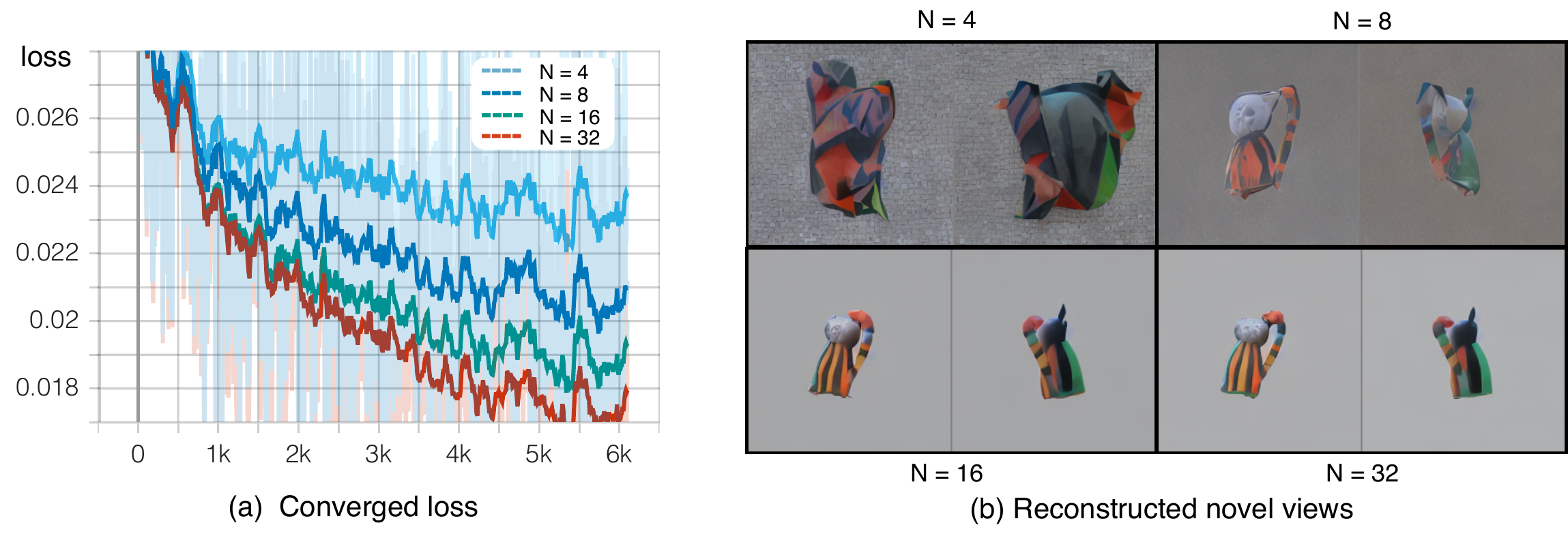}
    \caption{3D reconstructed results when aligning 3D representation and text embedding under different embedding sizes N.}
    \label{fig:ablation}
\end{figure*}

\begin{figure*}[t]
    \centering \includegraphics[width=0.98\textwidth]{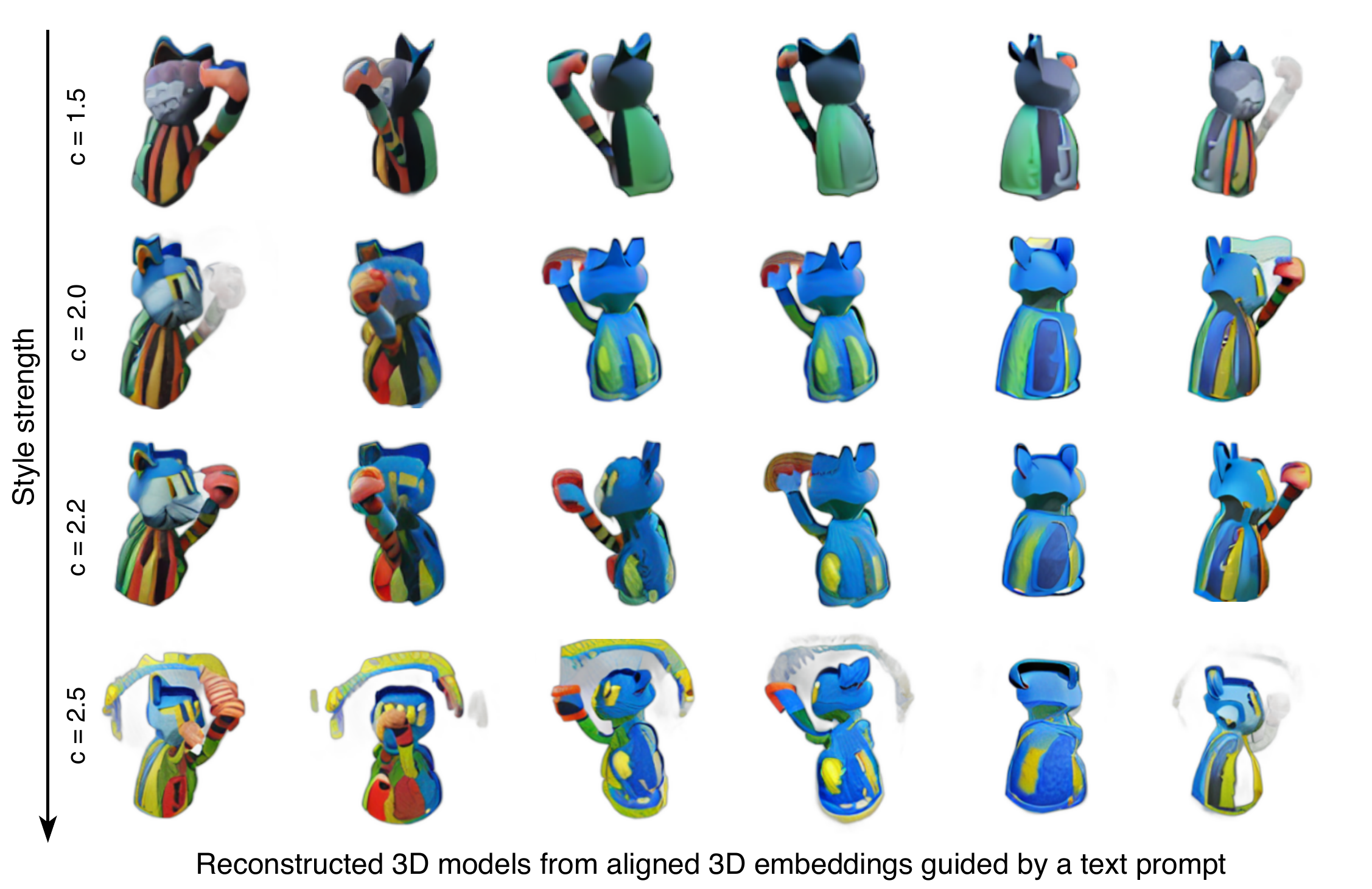}
    \vspace{-0.4cm}
    \caption{Visualized results of 3D personalized reconstruction from aligned 3D embeddings $S_{*}$. These view images are generated with a guided text prompt ``$\texttt{with Van Gogh style}$'' controlled by \cref{eq:prompt}. *Please note that the grayed areas are caused by alpha skinning~\cite{porter1984compositing}. }
    \label{fig:main_results_personalize}
\end{figure*}

\section{Experiments}
\noindent \textbf{Settings.} We experiment on AIGC-generated 3DGS models~\cite{kerbl2023_3dgs} to demonstrate the performance of our method. Specifically, we use images\footnote{\url{https://github.com/VAST-AI-Research/TriplaneGaussian}}\footnote{\url{https://github.com/3DTopia/LGM}} from TGS~\cite{zou2024triplane} and LGM~\cite{tang2024lgm} for generating 3DGS models. Large Multi-View Gaussian Model (LGM)~\cite{tang2024lgm} is adopted to reconstruct 3DGS models from the images.  To evaluate the reconstruction quality of our method, we asses the generated novel images via obtained 3D embeddings with rendered views from the 3DGS models (Sec.~\ref{sec:3d-text}). We also evaluate the impact of 3D embedding size (\textit{i.e.}, the number of embedding vectors used) on reconstruction quality (Sec.~\ref{sec:ablation}). To demonstrate downstream application with aligned 3D embedding, we conduct personalized 3D generation. The obtained 3D embeddings with text prompts are used to guide personalized 3D generation (Sec.~\ref{sec:3d-text}). Additionally, to further demonstrate the potential application of our method, we also align 2D images into a 3D embedding space for personalized 3D generation (Sec.~\ref{sec:2d-text}).

\subsection{Align 2D image and text embedding} 
\label{sec:2d-text}
Preliminary validation experiments are conducted by aligning 2D images and text embedding via text inversion~\cite{galimage_text_inversion}. As shown in \cref{fig:single_text}, the reconstructed 3D content still maintains the characteristics of the original samples. The experimental results demonstrate that the inverted text embedding can partially represent the features of the original image as the semantics embedding is aligned with the object's visual feature. We adopt this semantics embedding into a text-to-3D model for personalized 3D reconstruction. This approach enables stylized generation and achieves a certain degree of object reconstruction. However, such an initial attempt can not represent the 3D scene with the obtained embeddings, as 2D semantics alignment underscores the profound complexity of establishing equivalent vision-language bridges for 3D representations.

\subsection{3D reconstruction from inverted embedding}
\label{sec:3d-text}
To verify the effectiveness of our framework, we conduct an experiment of 3D reconstruction from the 3D embedding. As shown in \cref{fig:3d_rec}, the recovered 3D content across diverse cases from inverted embedding inputs, confirming the feasibility of this inverse mapping. This work provides the first empirical demonstration that 3D representations can be reconstructed directly from text embeddings in vision-language space. Our framework successfully transforms optimized embeddings $\mathbf{z}^*$ back into 3D content through a feedforward generation process~\cite{shi_mvdream}. This fundamentally validates that vision-language embeddings can encode sufficient geometric information for 3D reconstruction, establishing a new paradigm for content creation.

\subsection{3D content personalization from inverted embedding}
We demonstrate how our aligned text-3D embeddings enable natural language control over 3D content. 
We apply semantic vector arithmetic within CLIP space~\cite{radford2021learning_clip}, defined as $\mathbf{z}_{\text{edit}} = \mathbf{z}^* + \Delta \mathbf{z}_{\text{text}}$, and utilize the attention re-weighting~\cite{hertzprompt-to-prompt} to refine personalizing conditions by increasing the weighting factor for $\mathbf{z}_{\text{text}}$ in the attention mechanism. As demonstrated in \cref{fig:main_results_personalize}, we achieve consistent attribute modification in style transformation (``Van Gogh style''). More importantly, these transformations required no retraining or reoptimization of 3D parameters, demonstrating that scene manipulation can occur entirely within the embedding space.

\subsection{Ablation study}
\label{sec:ablation}
To further investigate the contributions of various components of our framework, we conduct an ablation study focused specifically on the size of the 3D embeddings used for reconstruction. We test different embedding sizes to assess their impact on reconstruction quality and loss convergence. As shown in \cref{fig:ablation}, our experiments reveal that an embedding size of \(N = 32\) achieves satisfactory results in terms of both reconstruction quality and convergence behavior. In contrast, when the embedding size is relatively small, effective reconstruction proves challenging, indicating that insufficient embedding capacity hinders the model’s ability to capture the necessary details for accurate 3D representation. The size of 3D embeddings in our proposed framework can be selected and adjusted for effective 3D model reconstruction.

%% file: sec/5_conclusion.tex
\section{Conclusion}
This paper presents a method for inverting 3D representations into a text-aligned embedding space and demonstrates its potential for both 3D reconstruction and personalized generation. By introducing a camera-conditioned 3D-to-text inversion and coupling it with attention re-weighting during generation, our framework establishes a direct semantic bridge between 3D representations and natural language, enabling intuitive, prompt-based control while reducing reliance on costly multi-view distillation and repeated retraining. Empirically, we show that the learned 3D embeddings can drive multi-view consistent synthesis and support flexible, language-guided personalization via simple operations in the embedding space, indicating that text-aligned embeddings can capture essential 3D structure and appearance while preserving editability. Beyond improving accessibility for creators, this embedding-centric paradigm suggests a path toward reusable and compositional 3D assets whose behavior can be specified and adapted through text.

\noindent \textbf{Limitations and future work.}
Although our method demonstrates the benefits of aligning 3D and text, this study provides only preliminary evidence on a limited set of scenarios and relies on pretrained backbones with per-scene optimization. To our knowledge, no existing approach can directly encode 3D representations into text embeddings; our method offers one practical route to obtain 3D embeddings and illustrate their uses. Future work should pursue direct, amortized 3D to text encoders for NeRF and 3DGS, train with paired or synthetic 3D and text data, and design metrics that quantify alignment and multiview consistency. We will also extend our 3D-to-text inversion to scene-level editing, which can enable prompt-guided manipulation such as occlusion removal and reflection removal across viewpoints. These steps can move the field toward tighter integration between 3D representations and natural language workflows.

\section*{Acknowledgment}
Renjie Group is supported by the National Natural Science Foundation of China under Grant No. 62302415, Guangdong Basic and Applied Basic Research Foundation under Grant No. 2022A1515110692, 2024A1515012822.